\pdfoutput=1

\documentclass[11pt]{article}

\usepackage{acl}

\usepackage{times}
\usepackage{latexsym}

\usepackage[T1]{fontenc}

\usepackage[utf8]{inputenc}

\usepackage{microtype}

\usepackage{hyperref} 
\usepackage{natbib} 
\usepackage{multirow} 
\usepackage{booktabs} 
\usepackage{tabularx} 
\usepackage{amsmath} 
\usepackage{amssymb} 
\usepackage[ruled,linesnumbered]{algorithm2e} 
\usepackage{xspace} 
\usepackage{pgfplots}\pgfplotsset{compat=1.9} 
\usepackage{balance} 
\usepackage{soul} 
\usepackage{subcaption} 
\usepackage{bm} 
\usepackage{mathtools} 

\usepackage{xcolor}
\newcommand{\new}[1]{#1}

\makeatletter
\newcommand*{\transpose}{\bgroup\@ifstar{\mathpalette\@transpose{\mkern-3.5mu}\egroup}{\mathpalette\@transpose\relax\egroup}}
\newcommand*{\@transpose}[2]{\setbox0=\hbox{\m@th$#1#2\intercal$}\raise\dp0\box0}
\makeatother

\DeclarePairedDelimiter{\norm}{\lVert}{\rVert} 

\title{Temporal Attention for Language Models}

\author{
    Guy D. Rosin \and Kira Radinsky \\
    Technion -- Israel Institute of Technology, Haifa, Israel \\
    \texttt{\{guyrosin,kirar\}@cs.technion.ac.il}
}

\begin{document}
\maketitle
\begin{abstract}

Pretrained language models based on the transformer architecture have shown great success in NLP.
Textual training data often comes from the web and is thus tagged with time-specific information, but most language models ignore this information.
They are trained on the textual data alone, limiting their ability to generalize temporally.
In this work, we extend the key component of the transformer architecture, i.e., the self-attention mechanism, and propose temporal attention---a time-aware self-attention mechanism.
Temporal attention can be applied to any transformer model and requires the input texts to be accompanied with their relevant time points. 
It allows the transformer to capture this temporal information and create time-specific contextualized word representations.
We leverage these representations for the task of semantic change detection; we apply our proposed mechanism to BERT and experiment on three datasets in different languages (English, German, and Latin) that also vary in time, size, and genre.
Our proposed model achieves state-of-the-art results on all the datasets.

\end{abstract}

\section{Introduction}
\label{sec:intro}

Language models (LMs) are usually pretrained on corpora derived from a snapshot of the web crawled at a specific moment in time~\cite{devlin2019bert,liu2019roberta}. 
But our language is constantly evolving; new words are created, meanings and word usages change.
For instance, the COVID-19 pandemic has caused significant changes to our language; consider the new video-related sense of ``Zoom'' and the new senses recently associated with the word ``vaccine''. 

\new{The ``static'' nature of existing LMs makes them unaware of time, and in particular unware of language changes that occur over time. This prevents such models from adapting to time and generalizing temporally~\cite{rottger2021temporal,lazaridou2021mind,hombaiah2021dynamic,dhingra2022time,agarwal2021temporal,loureiro2022timelms}, abilities that were shown to be important for many tasks in NLP and Information Retrieval~\cite{kanhabua2016temporal,rosin2017learning,huang2019neural,rottger2021temporal,savov2021predicting}.}
Recently, to create time-aware models, the NLP community has started to use time as a feature in training and fine-tuning language models~\cite{dhingra2022time,rosin2022time}.
These two studies achieve this by concatenating a time token to the text sequence before training the models.
The former was concerned with temporal question answering, whereas the latter---with semantic change detection and sentence time prediction.
\new{In this work, we introduce a new methodology to create time-aware language models and experiment on the task of semantic change detection.}

At the heart of the transformer architecture is the self-attention mechanism~\cite{vaswani2017attention}.
This mechanism allows the transformer to capture the complex relationships between words by relating them to each other multiple times.
An attention weight has a clear meaning: how much a particular word will be weighted when computing the next representation for the current word~\cite{clark2019does}.
This mechanism also enables the above-mentioned temporal models~\cite{dhingra2022time,rosin2022time} to work; by concatenating time-specific tokens to the text sequences, the self-attention mechanism would compute the relationships between them and the original tokens in the texts, effectively making the output embeddings time-aware (as the output embeddings will depend on the concatenated time tokens).

\begin{figure}
\centering
\includegraphics[width=1\linewidth]{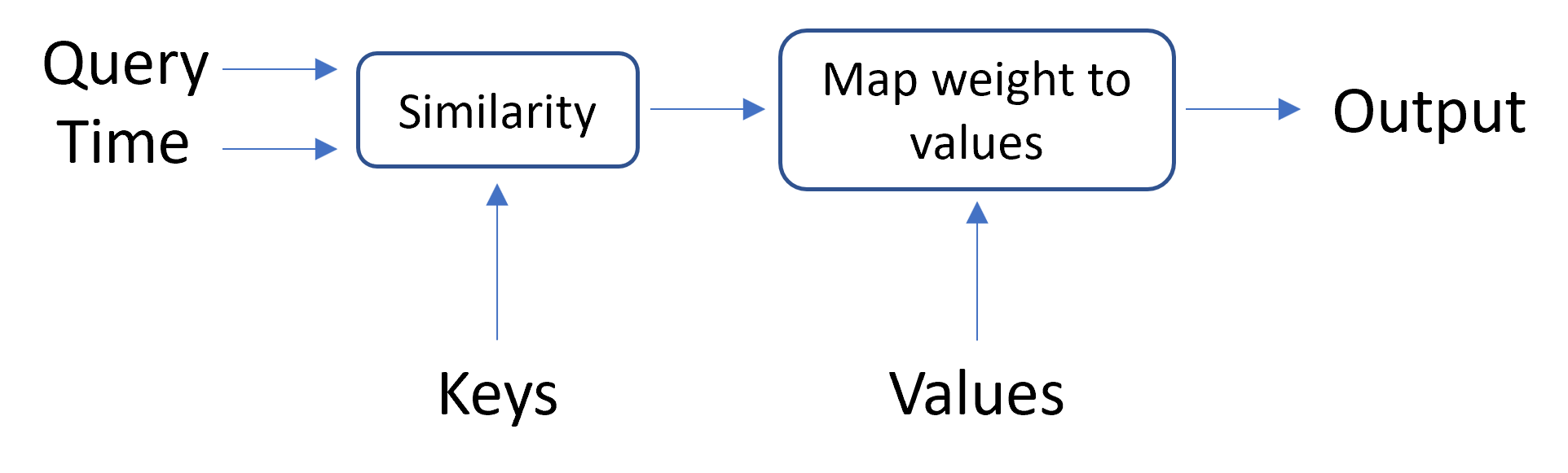}
\caption{\label{fig:temporal_attention_illustration}High-level illustration of our proposed temporal attention mechanism.}
\end{figure}

In this work, instead of changing the text sequences as in prior work, we modify the model itself and specifically the attention mechanism to make it time-aware. 
We propose a time-aware self-attention mechanism that is an extension of the self-attention mechanism of the transformer. It considers the time the text sequences (or documents) were written when computing attention scores.
As described above, self-attention captures relationships between words. 
We want to condition these relationships on time.
By adding a time matrix as an additional input to the self-attention (along with the standard query, key, and value matrices), we condition the attention weights on the time.
In other words, the adapted mechanism also considers the time when calculating the weights of each word.
We refer to this adapted attention as \textit{Temporal Attention} (Section~\ref{sec:temporal_attention}).
See Figure~\ref{fig:temporal_attention_illustration} for an illustration of our proposed mechanism.

We experiment on the task of semantic change detection --- the task of identifying which words undergo semantic changes and to what extent.
Semantic change detection methods are used in historical linguistics and digital humanities to study the evolution of word meaning over time and in different domains~\cite{kutuzov2018diachronic}.
Most existing contextual methods detect changes by first embedding the target words in each time point and then either aggregating them to create a time-specific embedding~\cite{martinc2020leveraging}, or computing a cluster of the embeddings for each time~\cite{giulianelli2020analysing,martinc2020capturing,montariol2021scalable,laicher2021explaining}. 
The embeddings or clusters are compared to estimate the degree of change between different times.
We experiment with several diverse datasets in terms of time, language, size, and genre. 
Our empirical results show that our model outperforms state-of-the-art methods~\cite{schlechtweg2019wind,martinc2020leveraging,montariol2021scalable,rosin2022time}.

Our contributions are threefold:
(1) We introduce a time-aware self-attention mechanism as an extension of the original mechanism of the transformer. 
The proposed mechanism considers the time the text sequences were written. The time is considered during the computation of attention scores, thus allowing to create time-specific contextualized word representations;
(2) We conduct evaluations on the task of semantic change detection and reach state-of-the-art performance on three diverse datasets in terms of time, language, size, and genre; 
(3) We contribute our code and trained models to the community for further research.\footnote{\url{https://github.com/guyrosin/temporal_attention}}

\section{Related Work}

\subsection{Temporal Language Models}
Several recent studies have explored and evaluated the generalization ability of language models to time~\cite{rottger2021temporal,lazaridou2021mind,agarwal2021temporal,hofmann2021dynamic,loureiro2022timelms}.
To better handle continuously evolving web content, \citet{hombaiah2021dynamic} performed incremental training.
\citet{dhingra2022time} experimented with temporal language models for question answering. 
They focused on temporally-scoped facts and showed that conditioning temporal language models on the temporal context of textual data improves memorization of facts.
\citet{rosin2022time} similarly concatenated time tokens to text sequences and introduced the concept of time masking (specific masking for the added time tokens). They focused on two temporal tasks: semantic change detection and sentence time prediction. 
Others focused on document classification by using word-level temporal embeddings~\cite{huang2019neural} and adapting pretrained BERT models to domain and time~\cite{rottger2021temporal}.
Recently, \citet{hofmann2021dynamic} jointly modeled temporal and social information by changing the architecture of BERT and connecting embeddings of adjacent time points via a latent Gaussian process.

In this work, we create a temporal LM by adapting the transformer's self-attention mechanism to time. The model receives each text sequence along with its writing time and uses both as input to the temporal attention mechanism. As a result, the model creates time-specific contextualized word embeddings.

\subsection{Semantic Change Detection}
Semantic change detection is the task of identifying words that change meaning over time~\cite{kutuzov2018diachronic,tahmasebi2018survey}. 
This task is often addressed using time-aware word representations that are learned from time-annotated corpora and then compared between different time points ~\cite{jatowt2014framework,kim2014temporal,kulkarni2015statistically,hamilton2016diachronic,dubossarsky2019time,del2019short}.
\citet{gonen2020simple} used a simple nearest-neighbors-based approach to detect semantically-changed words.
Others learned time-aware embeddings simultaneously over all time points to resolve the alignment problem, by regularization~\cite{yao2018dynamic}, modeling word usage as a function of time~\cite{rosenfeld2018deep}, Bayesian skip-gram~\cite{bamler2017dynamic}, or exponential family embeddings~\cite{rudolph2018dynamic}.

All aforementioned methods limit the representation of each word to a single meaning, ignoring the ambiguity in language and limiting their sensitivity.
Recent contextualized models (e.g., BERT~\cite{devlin2019bert}) overcome this limitation by taking sentential context into account when inferring word token representations.
Such models were applied to diachronic semantic change detection, where most detect changes by creating time-specific embeddings or computing a cluster of the embeddings for each time, and then comparing these embeddings or clusters to estimate the degree of change between different times~\cite{hu2019diachronic,martinc2020capturing,martinc2020leveraging,giulianelli2020analysing,laicher2021explaining,montariol2021scalable}.
Recently, \citet{rosin2022time} suggested another approach of detecting semantic change through predicting the writing time of sentences.
In our work, we use language models to create time-specific word representations and compare them to detect semantic change. While the above studies used language models as is, we modify their inner workings to make them time-aware by adapting the self-attention mechanism to time.

\section{Model}
\label{sec:model}

Our model adopts a multi-layer bidirectional transformer~\cite{vaswani2017attention}.
It treats words in the document as input tokens and computes a representation for each token. 
Formally, given a sequence of $n$ words $w_1, w_2, \dots, w_n$, the transformer computes $D$-dimensional word representations $x_1, x_2, \dots, x_n \in \mathbb{R}^D$.

\subsection{Self-Attention}
\label{sec:self_attention}

The self-attention mechanism is the foundation of the transformer~\cite{vaswani2017attention}.
It relates tokens to each other based on the attention score between each pair of tokens. 
In practice, the attention function is computed on a set of tokens simultaneously; our input sequence is packed together into a matrix \new{$X \in \mathbb{R}^{n \times D}$, in which each row $i$ corresponds to a word representation $x_i$ in the input sentence.}
We denote three trainable weight matrices by $W_Q, W_K, W_V \in \mathbb{R}^{D \times d_k}$.
We then create three distinct representations, i.e., query, key, and value:
$Q = XW_Q$, $K = XW_K$, $V = XW_V$, respectively, where $Q, K, V \in \mathbb{R}^{n \times d_k}$.

An attention function can be described as mapping a query and a set of key-value pairs to an output, where the query, keys, values, and outputs are all vectors. 
The output is computed as a weighted sum of the values, where the weight assigned to each value is determined by the dot product of the query with all the keys:

\begin{equation}
\textit{Attention}(Q, K, V) = \mathrm{softmax}\left(\frac{QK^\transpose}{\sqrt{d_k}}\right)V    
\end{equation}

\begin{figure}
\centering
\includegraphics[width=0.32\linewidth]{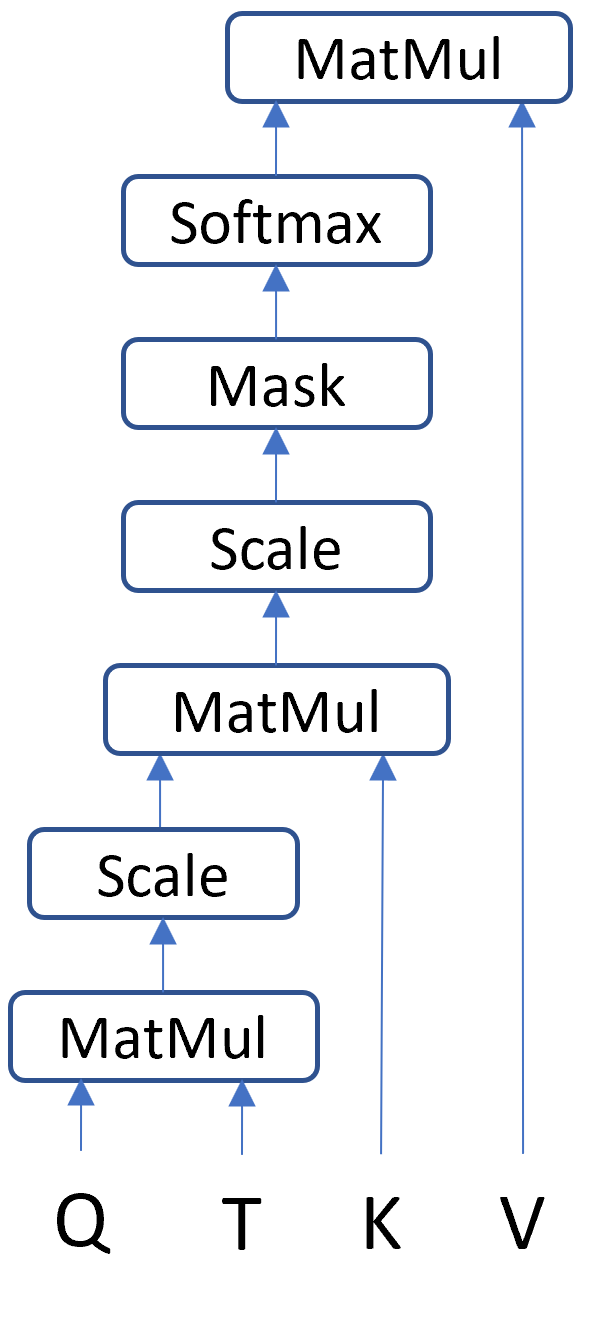}
\caption{\label{fig:temporal_attention}Illustration of our proposed temporal attention mechanism.}
\end{figure}

\subsection{Temporal Attention}
\label{sec:temporal_attention}

We now describe the temporal attention mechanism.
In the temporal setting, \new{similarly to the vocabulary of the model, our model has a vocabulary of time points. Theoretically, each token in an input sequence could have its own time point, but we simplify and assume the most common case where text sequences always refer to a single time point $t$.\footnote{Our mechanism also supports the setting where different tokens in a sequence are associated with different time points.}}
Given a sequence of $n$ words $w_1, w_2, \dots, w_n$ and its corresponding time point $t$, our model computes $D$-dimensional time-specific word representations $x_1^t, x_2^t, \dots, x_n^t$, where $x_i^t \in \mathbb{R}^D$.
As a by-product, we also compute $D$-dimensional time representations for the time points.
\new{Now, similarly to the input matrix $X$ (Section~\ref{sec:self_attention}), we define an embedding matrix $X^t \in \mathbb{R}^{n \times D}$ where each row $i$ contains the embedding vector of $x_i$'s time point.\footnote{Most tokens share the same time point, as noted above, except for special tokens such as padding and masking tokens, to which we associate unique time points.}}

To incorporate time in the attention mechanism, we use an additional trainable weight matrix $W_T \in \mathbb{R}^{D \times d_k}$ and create its corresponding representation matrix $T = X^t W_T$. 
Note $T \in \mathbb{R}^{n \times d_k}$, i.e., its dimensions are the same as the key, query, and value matrices.

To calculate the attention scores, we multiply the query matrix by the time matrix and then multiply by its transposed matrix, to keep the dimensions intact.
We then divide by the time matrix's norm, to avoid getting too large values. 
Formally, we define temporal attention by:

\begin{equation}
\begin{split}
\textit{TemporalAttention}(Q, K, V, T) = \\
\mathrm{softmax}\left(\frac{Q\frac{T^\transpose T}{\norm{T}}K^\transpose}{\sqrt{d_k}}\right)V
\end{split}
\end{equation}

Intuitively, by multiplying the query by the time, the attention weights are now conditioned on the time, i.e., they are time-dependent.

Temporal attention can be used together with other, existing temporal language models, such as \cite{rosin2022time,dhingra2022time}. In these two models, a time-specific token is prepended to each sentence.
In comparison to those methods, our approach does not require changing the input text, as it only modifies the attention mechanism of the language model. 
We further discuss and compare the two methods in Section~\ref{sec:theoretical_analysis}.


\new{The temporal attention mechanism requires each input text to be accompanied with a time point. There are no constraints on these time points, i.e., the mechanism is agnostic to the time granularity and the number of time points in the model.}

\subsection{Theoretical Analysis}
\label{sec:theoretical_analysis}
We now theoretically analyze the temporal attention mechanism more deeply and compare it to existing time concatenation methods~\cite{dhingra2022time,rosin2022time}.
We omit the scaling factor $\sqrt{d_k}$ for readability.

We denote the row vectors of the matrices $Q$, $K$, $V$, and $T$ by $q_i$, $k_i$, $v_i$, and $t_i$, respectively.
The attention head computes attention weights $\alpha$ between all pairs of words as softmax-normalized dot products between the query and key vectors:
\begin{equation}
    \alpha_{ij} = \mathrm{softmax}\left(q_i k_j^\transpose\right)
\end{equation}
where $i,j \in \{ 1, \dots, n \}$.

The output $y_i$ of the attention head is a weighted sum of the value vectors:
\begin{equation}
    y_i = \sum_{j=1}^{n}{\alpha_{ij} v_j}
     = \sum_{j=1}^{n}{\mathrm{softmax}\left(q_i k_j^\transpose\right) v_j}
\end{equation}
Baseline models, such as \citet{rosin2022time} and \citet{dhingra2022time}, prepend the text sequence with a time token at index 0, resulting in:
\begin{equation}
    y_i = \sum_{j=0}^{n}{\alpha_{ij} v_j}
     = \sum_{j=1}^{n}{\alpha_{ij} v_j} + \mathrm{softmax}\left(q_0 k_0^\transpose\right) v_0
\end{equation}
As we can see, by concatenating the time token, we add query, key, and value vectors for that token, i.e., a time component is added to the weighted sum.

In contrast, by using temporal attention, the attention weights become:
\begin{equation}
    \alpha_{ij} = \mathrm{softmax}\left(q_i \frac{t_i t_j^\transpose}{\norm{T}} k_j^\transpose\right)
\end{equation}
The $i$-th output vector $y_i$ is computed as:
\begin{equation}
    y_i = \sum_{j=1}^{n}{\alpha_{ij} v_j}
     = \sum_{j=1}^{n}{\mathrm{softmax}\left(q_i \frac{t_i t_j^\transpose}{\norm{T}} k_j^\transpose\right) v_j}
\end{equation}
Intuitively, we multiply by the vectors of time to scale the attention weight $\alpha_{ij}$ by time.
We observe two main differences between our proposed mechanism and prior work:
\begin{enumerate}
\item The time component is more tightly integrated in temporal attention: instead of just adding a time component to the weighted sum, in temporal attention the time component is multiplied by every component in the sum.
\item \new{Temporal attention requires learning an additional weight matrix} $W_T \in \mathbb{R}^{D \times d_k}$. In prior work, each input sequence is prepended with a time token, i.e., its length $n$ is increased by 1. 
As a result, the temporal attention mechanism consumes more memory (as it has additional $D \cdot d_k$ trainable parameters\footnote{When using the standard BERT-base architecture: $D \cdot d_k = 768 \cdot 64 = 49{,}152$}), whereas prior work requires more time to train (as its sequences are longer).
From our experiments, the overhead of both methods is negligible compared to the memory consumption and training time of BERT \new{(see analysis in Section~\ref{sec:model_size_analysis})}.
\end{enumerate}

\section{Semantic Change Detection}
\label{sec:scd}
In this section, we employ our proposed temporal attention mechanism (Section~\ref{sec:temporal_attention}) for the task of semantic change detection~\cite{kutuzov2018diachronic,tahmasebi2018survey}.
The ability to detect and quantify semantic changes is important to lexicography, linguistics, and is a basic component in many NLP tasks. For example, search in temporal corpora, historical sentiment analysis, and understanding historical documents.
The objective of this task is to rank a set of target words according to their degree of semantic change between two time points $t_1$ and $t_2$.
In this work, we follow the practice of \cite{martinc2020leveraging,rosin2022time} to estimate the semantic change a word underwent and rank the target words based on these estimates.

Given a target word $w$, we generate time-specific representations of it and compare them to detect semantic changes. 
Algorithm~\ref{alg:scd} formally describes the method.
We begin by sampling $n$ sentences containing $w$ from each time point $t \in \{t_1, t_2\}$ (line 3).
For each sentence $sent$, we create a sequence embedding by running it through the temporal attention model (note the model receives as input both $sent$ and $t$) and extracting the model's hidden layers that correspond to $w$ (lines 5--6). 
We then choose the last $h$ hidden layers and average them to get a single vector (line 7).
This is the contextual word embedding of $w$, denoted by $v$. 
Following, the resulting embeddings are aggregated at the token level and averaged (line 10), in order to create a non-contextual time-specific representation for $w$ for each time $t$, denoted by $x_t$.
Finally, we estimate the semantic change of $w$ by measuring the cosine distance (\textit{cos\_dist}) between two time-specific representations of the same token (line 12).

\begin{algorithm}
	\DontPrintSemicolon
	\KwIn{$w$ (target word)}
	\KwIn{$t_1$ (first time point)}
	\KwIn{$t_2$ (last time point)}
	\KwIn{$C$ (diachronic corpus)}
	\KwIn{$n$ (\# of sentences to sample)}
	\KwIn{$h$ (\# of last hidden layers to extract)}
	\For{$t \in \{t_1, t_2\}$}{
	    $L_t \gets \{\}$\;
    	$S_w \gets n$ sentences sampled from $C(t, w)$\;
    	\For{$sent \in S_w$}{
        	$H \gets \textit{TempAttModel}(sent, t)$\;
        	$H_w \gets H[w]$\;
        	$v \gets \textit{AvgHiddenLayers}(H_w, h)$\;
        	$L_t.\textit{insert}(v)$\;
        }
    	$x_t \gets \textit{avg}(L_t)$\;
    }
	${score} = \textit{cos\_dist}(x_{t_1}, x_{t_2})$\;
	\KwRet{${score}$}
	\caption{Semantic change estimation}
	\label{alg:scd}
\end{algorithm}

\section{Experimental Setup}
\label{sec:setup}

\subsection{Data}
\label{sec:data}

\begin{table*}
\small
\centering
\begin{tabular}{llllllll}
    \toprule
    Corpus & C1 Source & C1 Time & C1 Tokens & C2 Source & C2 Time & C2 Tokens & Target Words\\
    \midrule
    SemEval-English & CCOHA & 1810--1860 & 6.5M & CCOHA & 1960--2010 & 6.7M & 37\\
    SemEval-Latin & LatinISE & -200--0 & 1.7M & LatinISE & 0--2000 & 9.4M & 40\\
    SemEval-German & DTA & 1800--1899 & 70.2M & BZ, ND & 1946--1990 & 72.4M & 48\\
    \bottomrule
\end{tabular}
\caption{\label{tab:scd_datasets}Corpora for semantic change detection. Each corpus is split into two time points, denoted by C1 and C2.}
\end{table*}

To train and evaluate our models, we use data from the SemEval-2020 Task 1 on Unsupervised Detection of Lexical Semantic Change~\cite{schlechtweg2020semeval}.
We use corpora provided by this task for English, German, and Latin, covering a variety of genres, times, languages, and sizes.
They are all long-term: the English and German corpora span two centuries each, and the Latin corpus spans more than 2000 years.
The German corpus is much larger than the other two (x7 -- x10).
Each corpus is genre-balanced, and split into two time points; see Table~\ref{tab:scd_datasets} for their statistics.

Each corpus is accompanied with labeled data for semantic change evaluation. 
We use the data from Subtask 2 of this task, where the objective is to rank a set of target words according to their degree of semantic change between $t_1$ and $t_2$. 
The provided data is a set of target words that are either words that changed their meaning(s) (lost or gained a sense) between the two time points, or stable words that did not change their meaning during that time.
The target words are balanced for part of speech (POS) and frequency.
Each target word was assigned a graded label (between 0 and 1) according to their degree of semantic change (0 means no change, 1 means total change).
For the English dataset, we follow \cite{montariol2021scalable} and remove POS tags from both the corpus and the evaluation set.

\subsection{Baseline Methods}
We use the following baseline methods:
\begin{enumerate}
    \item \citet{schlechtweg2019wind} train Skip-gram with Negative Sampling (SGNS) on two time points independently and align the resulting embeddings using Orthogonal Procrustes. They compute the semantic change scores using cosine distance.
    \item \citet{gonen2020simple} use SGNS embeddings as well. They represent a word in a time point by its top nearest neighbors according to cosine distance. Then, they measure semantic change as the size of intersection between the nearest neighbors lists in the two time points.
    \item \citet{martinc2020leveraging} were one of the first to use BERT for semantic change detection. They create time-specific embeddings of words by averaging token embeddings over sentences in each time point, and then compare them by calculating cosine distance.
    \item \citet{montariol2021scalable} use BERT to create a set of contextual embeddings for each word. They cluster these embeddings and then compare the cluster distributions across time slices using various distance measures. 
    We use their best-performing method for each dataset as reported in the paper, which uses affinity propagation for clustering word embeddings and either Wasserstein or Jensen-Shannon distance as a distance measure between clusters.
    \item \citet{rosin2022time} create a time-aware BERT model by preprocessing input texts to concatenate time-specific tokens to them, and then masking these tokens while training. They introduce two methods to measure semantic change, namely temporal-cosine and time-diff. We use their best-performing method as reported in the paper, which is temporal-cosine.
    \item ``Scaled attention'': We present several baselines which are simplified versions of our temporal attention mechanism.
    Intuitively, our mechanism differentiates between different time points by learning a scaling factor per each pair of time points (based on the multiplication of learned time vectors; see Section~\ref{sec:temporal_attention}). 
    In these baselines, we use a constant scaling factor per time point and calculate attention weights using the following formula: 
    \begin{align*}
        \alpha_{ij} = \mathrm{softmax}\left(q_i s_{ij} k_j^\transpose\right)
    \end{align*}
    where $s_{ij}$ is the scaling factor.
    This scaling method can be seen as a combination of~\cite{martinc2020leveraging} and our temporal attention method.
    We present three options for $s_{ij}$: (1) Linear scaling. We hypothesize that recent texts should be given more weight, and define $s_{ij} = index(t_i)$, where $index(t_i)$ is the index of the time point $t_i$ out of all time points $t_1, \ldots, t_{n_t}$.
    (2) Exponential scaling: similarly to linear scaling, but using an exponent: $s_{ij} = 2^{index(t_i)}$.
    (3) Proportional to the number of documents: here we hypothesize that larger corpora should be given more weight, and define $s_{ij} = \frac{doc\_count(t_i)}{\sum_{k=1}^{n_t}{doc\_count(t_k)}}$, where $doc\_count(t_i)$ is the number of documents in $t_i$.
\end{enumerate}

\subsection{Our Method}

To train our models, for each language we use a pretrained BERT~\cite{devlin2019bert} model (bert-base-uncased, with 12 layers, 768 hidden dimensions, and 110M parameters) and post-pretrain it on the temporal corpus using our proposed temporal attention, as described in Section~\ref{sec:temporal_attention}. 
For semantic change detection, we use the method described in Section~\ref{sec:scd}.
We use the Hugging Face's Transformers library\footnote{\url{https://github.com/huggingface/transformers}} for our implementation.

Before training, we add any missing target words to the model's vocabulary. Since a pretrained model's vocabulary may not contain all the target words in our evaluation dataset, this is necessary to avoid the tokenizer splitting any occurrences of the target words into subwords \new{ (which we found out to reduce performance). The added words are randomly initialized.}

\subsection{Metrics}
We measure semantic change detection performance by the correlation between the semantic shift index (i.e., the ground truth) and the model's semantic shift assessment for each word in the evaluation set. 
We follow prior work~\cite{rosin2022time} and use both Pearson's correlation coefficient $r$ and Spearman's rank correlation coefficient $\rho$. 
The difference between them is that Spearman's $\rho$ considers only the ranking order, while Pearson's $r$ considers the actual predicted values.
In our evaluation, we make an effort to evaluate our methods and the baselines using both correlation coefficients, to make the evaluation as comprehensive as possible. 
There were some cases where we could not reproduce the original authors' results; in such cases, we opted to report only the original result.

\subsection{Implementation Details}
\label{sec:impl}
Due to limited computational resources, we follow \citet{rosin2022time} and train our models with a maximum input sequence length of 128 tokens. 
We perform all experiments on a single NVIDIA Quadro RTX 6000 GPU. 
We tune the following hyperparameters for each language: for training: learning rate in $\{1\text{e-}8, 1\text{e-}7, 1\text{e-}6, 1\text{e-}5, 1\text{e-}4\}$ and number of epochs in $\{1, 2, 3, 4\}$.
For inference: number of last hidden layers to use for embedding extraction $h \in \{1, 2, 4, 12\}$.

The chosen pretrained model and hyperparameters, along with the steps number and training time per language are as follows:

\begin{itemize}
\item For English: bert-base-uncased,\footnote{\url{https://huggingface.co/bert-base-uncased}} with $1\text{e-}9$ learning rate for 2 epochs (6.3K steps, took 70 minutes); all (12) hidden layers for inference.
\item For Latin: latin-bert,\footnote{\url{https://github.com/dbamman/latin-bert}} with $1\text{e-}5$ learning rate for 1 epoch (3.5K steps, took 25 minutes); last hidden layer for inference.
\item For German: bert-base-german-cased,\footnote{\url{https://huggingface.co/bert-base-german-cased}} with $1\text{e-}6$ learning rate for 1 epoch (38.1K steps, took 10 hours); last hidden layer for inference.
\end{itemize}

\begin{table*}
\centering
\begin{tabular}{lcccccc}
    \toprule
    \multirow{2}[3]{*}{Method} & \multicolumn{2}{c}{SemEval-Eng} & \multicolumn{2}{c}{SemEval-Lat} & \multicolumn{2}{c}{SemEval-Ger}\\
    \cmidrule(lr){2-3} \cmidrule(lr){4-5} \cmidrule(lr){6-7} &
    $r$ & $\rho$ & $r$ & $\rho$ & $r$ & $\rho$\\
    \midrule
    \citet{schlechtweg2019wind} & 0.512 & 0.321 & 0.458 & 0.372 & -- & 0.712\\
    \citet{gonen2020simple} & 0.504 & 0.277 & 0.417 & 0.273 & -- & 0.627\\
    \citet{martinc2020leveraging} & -- & 0.315 & -- & 0.496 & -- & 0.565\\
    \citet{montariol2021scalable} & 0.566 & 0.456 & -- & 0.488 & 0.618 & 0.583\\
    \citet{rosin2022time} & 0.538 & 0.467 & 0.485 & 0.512 & 0.592 & 0.582\\
    Scaled Linear Attention & 0.517 & 0.506 & 0.524 & 0.478 & 0.580 & 0.550\\
    Scaled Exp. Attention & 0.491 & 0.487 & 0.633 & 0.528 & 0.569 & 0.526\\
    Scaled by Doc Attention & 0.532 & 0.478 & 0.657 & 0.505 & 0.595 & 0.567\\
    Temporal Attention & \textbf{0.620} & \textbf{0.520} & \textbf{0.661} & \textbf{0.565} & \textbf{0.767} & \textbf{0.763}\\
    \bottomrule
\end{tabular}
\caption{\label{tab:scd}Semantic change detection results on SemEval-English, SemEval-Latin, and SemEval-German, measured using Pearson's $r$ and Spearman's $\rho$ correlation coefficients.}
\end{table*}

\section{Results}
\label{sec:results}
In this section, we outline the results of our empirical evaluation.
In all tables throughout the section, the best result in each column is boldfaced; performance is measured using Pearson's $r$ and Spearman's $\rho$ correlation coefficients.

\subsection{Main Result}
Table~\ref{tab:scd} shows the results for semantic change detection on the SemEval datasets.
Our temporal attention model outperforms all the baselines for all datasets and metrics with significant correlations ($p<0.0005$) and large margins (7\%--36\%).
We observe moderate to strong correlations (around 0.52--0.76) for all datasets.
Even for the German dataset, on which recent BERT-based methods got relatively lower results (and were outperformed by word2vec-based methods such as \citet{schlechtweg2019wind}), our model achieves strong correlations and state-of-the-art performance.
In Section~\ref{sec:analysis_attention} and Section~\ref{sec:model_size_analysis}, we experiment with variations of our method and achieve even stronger performance on the English dataset.

Finally, looking at the three scaled attention baselines, they all perform similarly and are positioned between \citet{martinc2020leveraging} and our temporal attention model, as expected.

\subsection{Temporal Attention with Temporal Prepend}
\label{sec:analysis_attention}

Until now, we used temporal attention on BERT~\cite{devlin2019bert} to create our model.
In this section, in addition to using temporal attention, we also prepend a time token to the input sequences, as done in~\citet{rosin2022time}. 
That is, we experiment with applying temporal attention on top of their model.

\new{Table~\ref{tab:analysis_attention} shows the results of this combined model compared to each of its components.
First, prepending time tokens is inferior to the other models.
When comparing our proposed temporal attention and the combined model, we observe mixed results: temporal attention alone works better for the Latin and German datasets, but for the English dataset the combination of temporal attention and prepending time tokens performs better.}

\begin{table}
\footnotesize
\centering
\setlength{\tabcolsep}{0.42em}
\begin{tabular}{lcccccccc}
    \toprule
    \multirow{2}[3]{*}{Method} & \multicolumn{2}{c}{SE-Eng} & \multicolumn{2}{c}{SE-Lat} & \multicolumn{2}{c}{SE-Ger}\\
    \cmidrule(lr){2-3} \cmidrule(lr){4-5} \cmidrule(lr){6-7} & 
    $r$ & $\rho$ & $r$ & $\rho$ & $r$ & $\rho$\\
    \midrule
    Temp. Prep. & 0.538 & 0.467 & 0.485 & 0.512 & 0.592 & 0.582\\
    Temp. Att. & 0.620 & 0.520 & \textbf{0.556} & \textbf{0.556} & \textbf{0.767} & \textbf{0.763}\\
    Both & \textbf{0.655} & \textbf{0.548} & 0.541 & 0.508 & 0.645 & 0.682\\
    \bottomrule
\end{tabular}
\caption{\label{tab:analysis_attention}Semantic change detection results on the English, Latin, and German datasets, comparing \new{time token prepending~\cite{rosin2022time} with our proposed temporal attention, and a combination of both}.}
\end{table}

\subsection{Impact of BERT Model Size on Temporal Attention}
\label{sec:model_size_analysis}
Our model is based on the most commonly used pretrained BERT model, called BERT-base, which contains 12 transformer layers and a hidden dimension size of 768. 
In this section, we train and evaluate models of different sizes, namely `small' and `tiny', that are based on much smaller pretrained variants of BERT:
BERT-small\footnote{\url{https://huggingface.co/prajjwal1/bert-small}} has 26\% of the parameters of BERT-base, containing only 4 transformer layers while its hidden dimension is 512;
BERT-tiny\footnote{\url{https://huggingface.co/prajjwal1/bert-tiny}} has just 4\% of the parameters of BERT-base, with 2 transformer layers and a hidden dimension of 128.
We perform this evaluation only for the SemEval-English dataset, as smaller pretrained BERT models are currently publicly available only for the English language.

Table~\ref{tab:scd_model_size_comparison} shows the comparison results, where we compare the three variants of our temporal attention model, along with the two variants of~\citet{rosin2022time}. \new{We also denote the number of trainable parameters for each model (see the theoretical analysis in Section~\ref{sec:theoretical_analysis}).}
We observe a clear negative correlation between model size and performance (measured by both Pearson's $r$ and Spearman's $\rho$); the smaller the model, the better the performance.
\new{While this finding may sound counterintuitive, it is in line with \citet{rosin2022time}}, who hypothesized that to understand time there is no need to use extremely large models, and reported higher-than-expected performance for the tiny model. 
In their study, that model achieved a slightly lower performance compared to their standard (base) model, but still outperformed most baselines. 
Overall, this is an encouraging finding; smaller models mean faster training and inference times, as well as smaller memory footprints. This lowers the bar to enter the field. 

\begin{table}
\small
\centering
\begin{tabular}{lccc}
    \toprule
    Method & \new{Params} & $r$ & $\rho$\\
    \midrule
    \citet{rosin2022time} base & 109.52M & 0.538 & 0.467\\
    \citet{rosin2022time} tiny & 4.42M & 0.534 & 0.427\\
    \addlinespace
    Temp. Att. base & 116.61M & 0.620 & 0.520\\
    Temp. Att. small & 29.85M & 0.660 & 0.584\\
    Temp. Att. tiny & 4.45M & \textbf{0.703} & \textbf{0.627}\\
    \bottomrule
\end{tabular}
\caption{\label{tab:scd_model_size_comparison}Results for semantic change detection for models of different sizes on SemEval-English.}
\end{table}

\subsection{Qualitative Analysis}
\begin{figure}
\centering
\includegraphics[width=\linewidth]{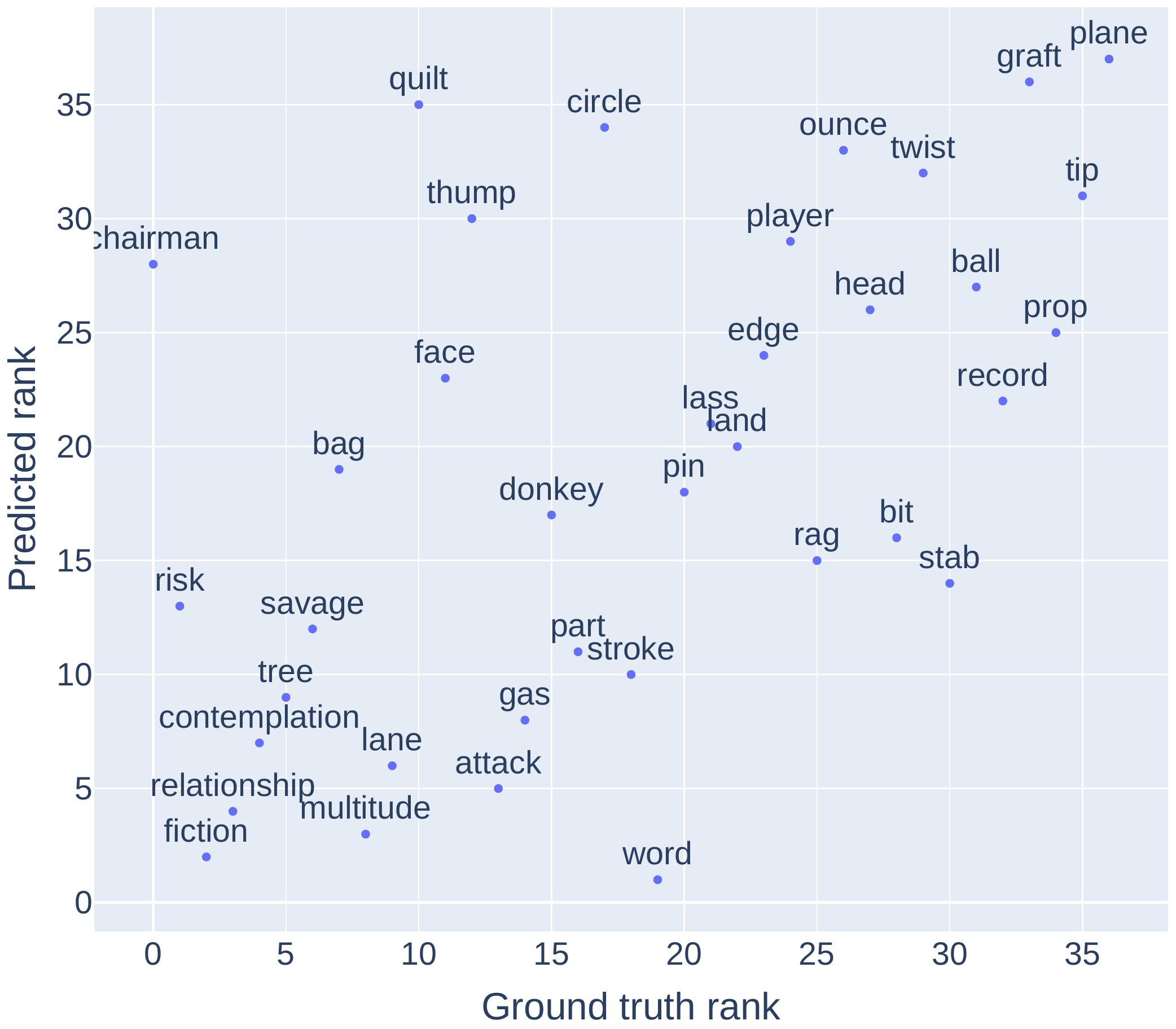}
\caption{\label{fig:semeval_eng_correlations}Semantic change detection on the SemEval-English dataset: ground truth ranks vs. our model's ranks (Spearman's $\rho = 0.520$).}
\end{figure}

Figure~\ref{fig:semeval_eng_correlations} shows the Spearman correlation between the ground truth ranks and our model's ranks for the SemEval-English dataset. 
The correlation is moderate ($0.520$), and we observe a similar number of false-positive words (top-left corner) and false negatives (bottom-right corner). 
Interestingly, we can see that the model performs better on the more changed words (right half, rank above 19, e.g., ``plane'', ``tip'', and ``head''), while there are more errors on the static words (left half, e.g., ``chairman'', ``risk'', and ``quilt'').
Most of the false positives seem to be either slang words or concerning word usages that are less likely to appear in our corpora which is mainly composed of newsletters and books (Section~\ref{sec:data}). 
For example, the verb ``stab'', which traditionally means to push a knife into someone, has a newer meaning of attempting to do something.
The noun ``word'' can be used to express agreement.

\section{Conclusion}
\label{sec:conclusions}
In this paper, we presented a time-aware self-attention mechanism as an extension of the original mechanism of the transformer. 
The proposed mechanism considers the time the text sequences were written when computing attention scores, thus allowing creating time-specific contextualized word representations.
We conducted evaluations on the task of semantic change detection and reached state-of-the-art performance on three diverse datasets in terms of time, language, size, and genre.
In addition, we experimented with small-sized pretrained models and found they outperform larger models on this task.
We conduct an experiment evaluating the marginal addition of time token prepending along with temporal attention and conclude that on all but the English dataset it hurts performance. We wish to study how to best combine the two approaches in future work.
Additionally, for future work, we plan to extend this work by applying temporal attention to other tasks, such as web search and sentence time prediction, as well as experimenting with more time points and different granularities.

\bibliography{anthology}
\bibliographystyle{acl_natbib}

\end{document}